\documentclass[runningheads]{llncs}
\usepackage{graphicx}

\usepackage{tikz}
\usepackage{comment}
\usepackage{amsmath,amssymb} 
\usepackage{color}
\usepackage{parskip}
\usepackage{caption}

\usepackage{textcomp}
\usepackage{hyperref}

\begin{document}

\pagestyle{headings}
\mainmatter
\def\ECCVSubNumber{100}  

\title{Multiple Generative Adversarial Networks Analysis for Predicting Photographers' Retouching \thanks{This project was supervised by Chen Liu for the Computational Photography (CS-413) course at EPFL.}} 

\titlerunning{Multiple GANs analysis for predicting photographer's retouching}

\author{Marc Bickel\inst{1}
\and
Samuel Dubuis\inst{1} 
\and
S\'ebastien Gachoud\inst{1}
}

\authorrunning{Bickel, Dubuis \& Gachoud}
\institute{\'Ecole Polytechnique F\'ed\'erale de Lausanne (EPFL), Switzerland\\
 School of Computer and Communication Sciences}

\maketitle

\footnotetext[1]{Author list in alphabetical order. Correspondance to :\\ \email{marc.bickel@epfl.ch}, \email{s.dubuis@epfl.ch}, \email{sebastien.gachoud@epfl.ch}}
\begin{abstract}
Anyone can take a photo, but not everybody has the ability to retouch their pictures and obtain result close to professional. Since it is not possible to ask experts to retouch thousands of pictures, we thought about teaching a piece of software how to reproduce the work of those said experts. This study aims to explore the possibility to use deep learning methods and more specifically, generative adversarial networks (GANs), to mimic artists' retouching and find which one of the studied models provides the best results.
\end{abstract}

\section{Introduction}
\vspace{-3mm}
Our goal was to recreate the artist's retouchings of the raw images with Deep Learning methods, especially using Generative Adversarial Networks (GANs) \cite{original_GAN}. 

Some non-deep learning methods fulfil that task but are very hard to use repeatedly. The retouchings done manually on one image do not translate well to other pictures, mainly because of image context and semantics. These retouchings are very subjective since they are the ``improvements" that this artist adds to the image. Different artists improve a given image in different ways. This makes repeating the transformations especially hard. Deep learning comes to the rescue, especially GANs, because they consider the original image, and apply the transformation more adaptively. This allows for the images to look more natural, closer to what a true artist would produce. We used different types of GANs to compare the results. GANs are the ideal type of neural networks for our project because the generator can learn the underlying distribution of the artistic preferences of an artist, and then apply this to a raw image, creating the desired output. 

Following recommendations, we focused our efforts on CycleGAN \cite{cyclegan}, but also looked around for other variants, such as StarGAN \cite{StarGAN}, a multi-domain variation of CycleGAN; StyleGAN, a style-transfer type GAN; Pix2Pix \cite{pix2pix}, a paired images style transfer GAN, and many others. Since we wanted transformations from image of the domain A (the RAW images) to image of the domain B (the chosen artist production), and these domains included all sorts of subjects (faces, landscapes, etc), we had to use a GAN with a sufficiently general architecture. CycleGAN allows for this, while not requiring paired images. This was for us an ideal setup that allowed us to jump rapidly into tuning the model. 

We propose a set of basic linear transformations applied to the raw image, before feeding it to a CycleGAN/Pix2Pix neural network. We believe this allows to produce near-human quality images, as if they were processed through an image editing software. Our work is based on the code from the CycleGAN/Pix2Pix paper \cite{cyclegan} and their GitHub repository\footnote{\url{https://github.com/junyanz/pytorch-CycleGAN-and-pix2pix}}.

We assume we have access to raw images, and that we know the visual tendencies of an artist, i.e. we have many examples of his work. A given GAN can only learn the distribution of a single artist, otherwise it would learn some mixture of preferences (StarGAN tries to address this). We present, in this paper, the architecture of the neural network we used, what pre-processing we put the raw images through, and the visual scores we obtained. The FID score is also given to numerically compare the outputs. 

\vspace{-3mm}
\section{Related work}
\vspace{-3mm}
\label{section:litterature}
We study two approaches to tackle the problem in this project: a GAN-based, or a lookup table based (LUT). The latter requires more comprehension of the domain and of the preferences of the artists, while the first needs more computational power. 
We think that the expressive power of GANs is higher than of LUTs, because LUTs would still require manual fine-tuning for every individual picture. 

\begin{figure}
    \centering
    \includegraphics[width=0.8\linewidth]{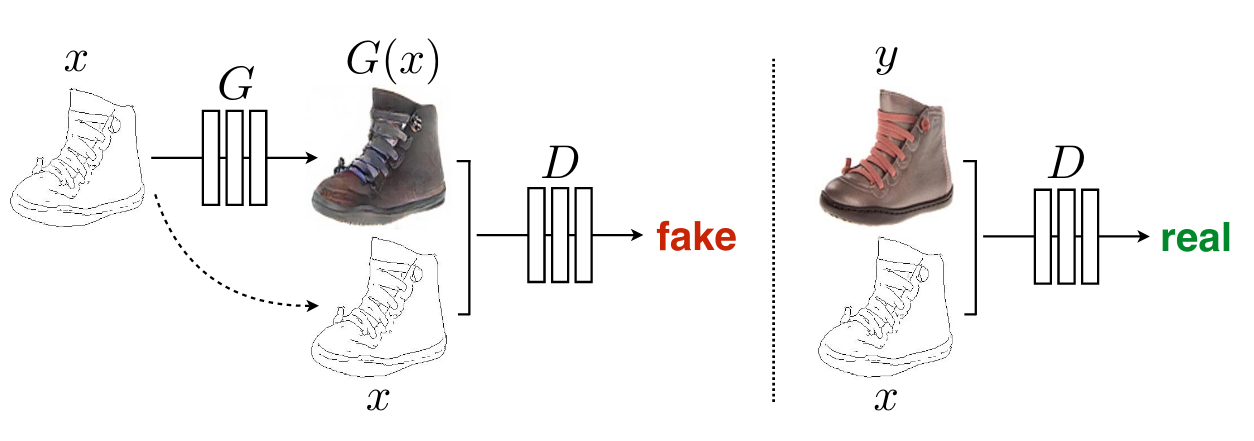}
    \caption{Taken from \cite{pix2pix}. Training a conditional GAN to map edges$\rightarrow$photo. The discriminator, $D$, learns to classify between fake (synthesized by the generator) and real {edge, photo} tuples. The generator, $G$, learns to fool the discriminator. Unlike an unconditional GAN, both the generator and discriminator observe the input edge map}
    \label{fig:pix2pix_scheme}
\vspace{-3mm}
\end{figure}

GANs were first introduced in \cite{original_GAN}, consisting of a generator and a discriminator. In details, GANs are generative models that learn a mapping from random noise vector $z$ to an output image $y$, $G:z \rightarrow y$ \cite{NIPS2014_5423}. In contrast, conditional GANs learn a mapping from observed image $x$ and random noise vector $z$, to an output image $y$, $G:{x,z} \rightarrow y$. The generator $G$ is trained to produce outputs that cannot be distinguished from ``real" images by an adversarially trained discriminator, $D$, which is trained to do as well as possible at detecting the generator’s ``fakes". This training procedure is diagrammed in Figure \ref{fig:pix2pix_scheme}, from \cite{pix2pix}. 

From this, different tweaks have been applied to the original idea, using different losses, different network architectures, while leaving the foundations intact. As said before, we thought this was the best approach because it is end-to-end, which is what we aim for. 

We then had a look at different GANs architectures and papers that would allow us to realise this project. Quickly, we were brought on to CycleGAN \cite{cyclegan}, which is a prime candidate for our project. It builds on the original GAN paper by not only working from the image domain A to image domain B, but also in the opposite direction, effectively creating a cycle, hence the name. However, we also considered other alternatives because we did not want to restrict ourselves to a single type of GAN.

The next GAN architecture studied was DiscoGAN \cite{DiscoGAN}. This GAN variation focuses on learning the cross-domain relations by using a reconstruction loss, given unpaired data. This approach seems to work well on faces, but the results for other images semantics such as landscapes looked worse than CycleGAN, so we did not pursue this architecture further. 

Next is StarGAN \cite{StarGAN}, which looked really promising, since it seems able to learn all the mappings at the same time. Our objective is to map each raw image to 5 corresponding retouched images, so the potential was there. While CycleGAN and many other architectures can learn the mapping from RAW to one artist ``distribution" at a time, StarGAN proposes to learn all of them at the same time. This showed great promise, therefore, we explored this in more detail in subsection \ref{subsection:Implementation:StarGAN}. A variant of StarGAN \cite{starGANv2} is proposed concurrently with this project. So we focus on the original starGAN here.

Pix2Pix \cite{pix2pix} is a more specialised version of CycleGAN. It requires paired images in the dataset, which it then uses to create a single X/Y image that holds both tightly correlated images. We explored that avenue as well, as explained in \ref{implementation:pix2pix}. 

Finally, we also tried to find an alternative to the generative power of GANs. Adversarial Latent Autoencoders \cite{styleALAE} are starting to catch up in terms of image quality. This architecture is very recent and allows for very impressive results. However, we found out that while the results on face images look impressive, other image semantics are subpar at best. We decided not to pursue this. 

After some testing, we chose CycleGAN as our main architecture, and the other would have to compete against it. This report does not hope to explain in detail what is CycleGAN. This website: \url{https://hardikbansal.github.io/CycleGANBlog/}, does a good job at this task, and we encourage readers to get familiar with this architecture.

\vspace{-3mm}
\section{Implementation}
\vspace{-3mm}
\subsection{Images Pre-processing}
\vspace{-3mm}
\label{subsection:preprocessing}
This project uses the \textit{fivek} dataset \cite{fivek}, consisting on 5000 raw pictures taken by photographers, and 25000 edited pictures by artists in the Adobe LightRoom\textregistered\ software, 5000 pictures per artist. 
The images from the \textit{fivek} dataset \cite{fivek} were too big to have any hope of processing them in a reasonable amount of time through any GAN. Therefore, the size of the images has been reduced such that the longest side would measure 500 pixels and the shortest would be kept having the closest possible ratio (short side / long side) to the original image. This method was chosen to comply with the practices followed by \cite{fivek}, which also are the baseline methods, explained further in section \ref{subsection:baseline-methods}. Later on the same procedure would be repeated by reducing long edges to 256 for processing in GANs. Trying different GAN models and tune their hyper-parameters does not require full size images, therefore, in order to reduce the time required to explore all models, the dataset has been further reduced to have the longest side to be 256 pixels long.

Every image reduction has been processed with a bicubic interpolation over 4×4 pixel neighborhood and the result has been saved in PNG format without compression.

The size reduction of the raw images could not be directly applied to the raw format. Therefore, before being resized, the raw images have been processed with the adaptive homogeneity-directed (AHD) demosaicing algorithm with full FDBB noise reduction, white balancing, pixel value scaling, 8 bits per sample and the output color space used was the Adobe color space. The Adobe color space has been chosen to match the color space of the images retouched by artists since they used Adobe Lightroom\textregistered\ as editing tool. The result has been saved as PNG without compression.
\vspace{-3mm}
\subsection{Baseline methods}
\vspace{-3mm}
\label{subsection:baseline-methods}
The baseline methods were created following the paper from Bychkovsky et al. \cite{fivek} and they are LUT based. Due to the lack of details about their theoretical part of the work, we have done comprehensive experiments here. 

Their idea was that the adjustment of the luminance curve of an image is responsible for the majority of the modification and variance made by artists on a raw image.

Thus we created a pipeline that would implement a part of their full work as our baseline methods. For each of the 30000 images, after a rescaling so that the longest edge measured 500 pixels, explained in subsection \ref{subsection:preprocessing}, we applied a transformation from the RGB channels to the LAB ones and normalized values between 0 and 100. 

Then we only focused on the first channel of our new images, the luminance values of each pixel, channel L. Following Bychkovsky et al. \cite{fivek} and after experimenting with these values, we created the curve of luminance by computing its cumulative distributive function (CDF). This curve was sampled 51 times, as our values went from 0 to 100, it means simply taking one out of two. 

After sampling, we had for each image 51 points, in ascending order, that represented the luminance CDF curve. Based on \cite{fivek}, we implemented a machine learning framework with, as input, the dataset of 51-sampled raw images and then each artist's 51-sampled edited images. We used a Gaussian Process Regressor (GPR) with 5-fold cross-validating.

Finally, we could input an image luminance 51-sampled CDF and have as output of this regressor the closest luminance curve to the one the artist would have made. What remains to do is a remapping of our pixels luminance value to approach the goal curve and then replace the L channel of the input image by the new one, and finally transform the image back to RGB.

\begin{figure}[h]
\begin{minipage}[c]{.6\linewidth}
    \centering
    \includegraphics[width=\linewidth]{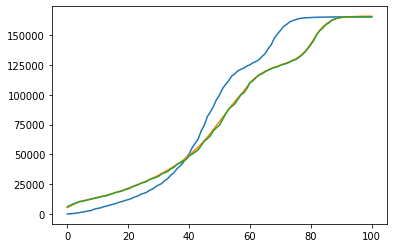}
\end{minipage}\hfill
\begin{minipage}[c]{.38\linewidth}
    \captionof{figure}{Example of the luminance CDF of the images referenced in figures \ref{fig:lum_raw} and \ref{fig:lum_gt}.\\[6pt]
    \small{The blue line represents the CDF of the raw image, seen in fig. \ref{fig:lum_raw}.\newline
    The orange line represents the goal CDF of the ground truth image, seen in fig \ref{fig:lum_gt}.\newline
    The green line represents the CDF after the mapping is done.
    }}
    \label{fig:cdf_example}
\end{minipage}
\vspace{-13mm}
\end{figure}

\begin{figure}[!h]
\begin{minipage}{.5\textwidth}
  \centering
  \captionsetup{width=.8\linewidth}
  \includegraphics[width=.7\linewidth]{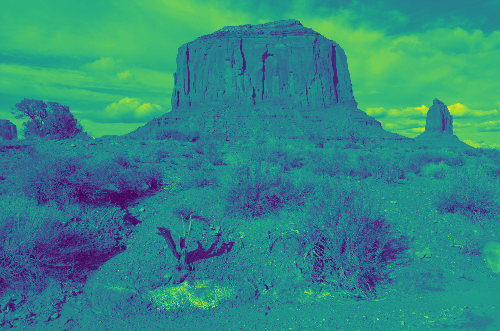}
  \captionof{figure}{Luminance values for each pixel of the image, from the raw image after post-processing.}
  \label{fig:lum_raw}
\end{minipage}
\begin{minipage}{.5\textwidth}
  \centering
  \captionsetup{width=.8\linewidth}
  \includegraphics[width=.7\linewidth]{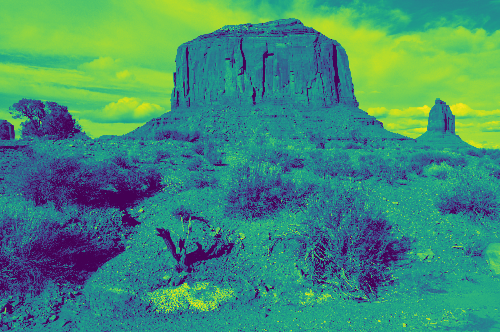}
  \captionof{figure}{Luminance values for each pixel of the image, from the ground truth image, for artist C.}
  \label{fig:lum_gt}
\end{minipage}
\end{figure}

As can be seen in Figure \ref{fig:cdf_example}, the two CDF of each image from figures \ref{fig:lum_raw} and \ref{fig:lum_gt} are disparate but our mapping did a great job in changing the pixels value.

\subsection{CycleGAN}
\label{subsection:CycleGAN}
\subsubsection{Model}

Our approach is based on CycleGAN \cite{cyclegan} and the associated code. The main contribution of that paper is the addition of a cycle-consistent loss, which enables the pairing of an inverse mapping $F : B \rightarrow A$ with the mapping $G : A \rightarrow B$ found in the classic GAN model (usually heavily under-constrained). This allows to enforce $F(G(A)) \approx A$ (and vice versa). The result is that training on unpaired data is a lot easier and produces better results. 

In practice, it means that the network learns the mapping from A to B at the same time as the mapping from B to A. Therefore, it needs two generators and two discriminators, making it quite a heavy GAN to run. We use two resnets 9 blocks as generators, and two PatchGANs, introduced by Pix2Pix \cite{pix2pix}, as discriminators.

\begin{figure}
    \centering
    \includegraphics[width=\textwidth]{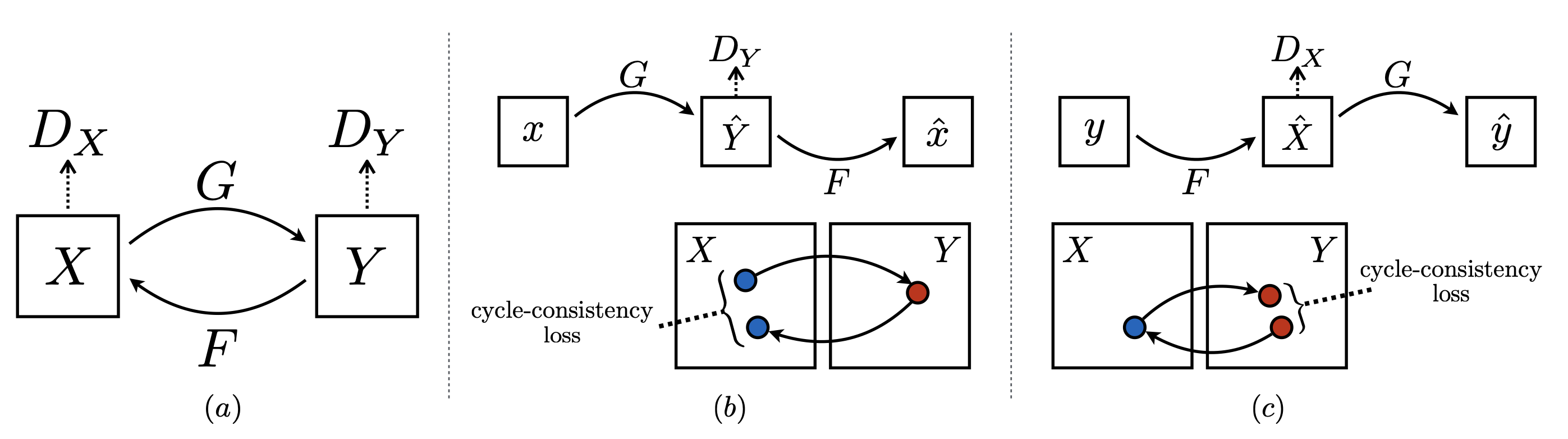}
    \caption{Figure taken from \cite{cyclegan}}
    \label{fig:cycle-consistency_loss}
\end{figure}

Figure \ref{fig:cycle-consistency_loss} shows that the model contains two mapping functions $G : X \rightarrow Y$ and $F : Y \rightarrow X$, and the associated adversarial discriminators $D_Y$ and $D_X$. $D_Y$ encourages $G$ to translate $X$ into outputs that are indistinguishable from domain $Y$, and vice versa for $D_X$ and $F$. To further regularize the mappings, \cite{cyclegan} introduces two cycle consistency losses that capture the intuition that if CycleGAN translates from one domain to the other and back again, it should arrive where it started: (b) forward cycle-consistency loss: $x \rightarrow G(x) \rightarrow F (G(x)) \approx x$, and (c) backward cycle-consistency loss: $y \rightarrow F (y) \rightarrow G(F (y)) \approx y$. 

The full objective function used is 
\begin{multline}
    \mathcal{L}(G, F, D_X, D_Y) = \\
    \mathcal{L}_{GAN}(G, D_Y, X, Y) + \mathcal{L}_{GAN}(F, D_X, Y, X) + \theta \mathcal{L}_{cyc}(G, F)
\end{multline}
where $\mathcal{L}_{GAN}$ is the adversarial loss, and $\mathcal{L}_{cyc}$ is the cyclic loss described before. 
\begin{multline}
    \mathcal{L}_{GAN}(G, D_Y, X, Y) = \\
    \mathbb{E}_{y\sim p_{data}(y)}[\log D_Y(y)] + \mathbb{E}_{x\sim p_{data}(x)}[\log(1-D_Y(G(x))]
\end{multline}
\begin{equation}
    \mathcal{L}_{cyc}(G, F) = 
    \mathbb{E}_{x\sim p_{data}(x)}[\|F(G(x))-x\|] + 
    \mathbb{E}_{y\sim p_{data}(y)}[\|G(F(y))-y\|_1]
\end{equation}
We encourage readers to look at \cite{cyclegan} for more details and mathematical justifications. 

\subsubsection{Our usage}

We found the code pretty satisfactory, hence we didn't change much regarding the architecture of CycleGAN itself. As mentioned earlier, we use the default Neural Networks architectures used by the authors: two resnets 9 blocks as generators, and two PatchGANs as discriminators. We did however change the hyperparameters of the model, in the scope of what was allowed by the authors and their code. We did fine-tune the following hyper-parameters: 

\begin{itemize}
    \itemsep-5pt    
    \item loss type
    \item learning rate
    \item learning rate decay policy
    \item norm type
    \item Adam optimizer momentum parameter
    \item pre-processing steps
\end{itemize}

By loss type, we mean what loss function is used for the adversarial loss. The default option is least-square version of the GAN objective function \cite{least-squares}. The implementation of CycleGAN trains in 2 phases. For the first 100 epochs, the learning rate stays the same. It does, however, decay over the 100 next epochs (there are 200 total epochs). The policy defines if the decay is done in steps, following some plateau, or another strategy. The frequency defines at what speed the decay happens. The norm type parameter allows to change between batch norm or instance norm. The momentum parameter of the Adam optimizer is explained in \cite{adam}. The pre-processing parameters refers to standard procedures when using convolutional neural networks to augment a dataset, i.e. randomly flip, rotate, and crop the images. These options are all included in the code of CycleGAN, so that we do not have to do it manually. 

For most of these hyper-parameters, the default values performed best. Since the authors certainly fine-tuned their model, this makes sense. One notable exception is present, however. The best-performing loss function is the vanilla GAN loss, presented in \cite{original_GAN}. It allows for better visual results.

In the end, we get the best results with the following configuration: 

\begin{itemize}
    \itemsep-5pt 
    \item loss type = vanilla loss
    \item learning rate = default
    \item learning rate decay policy = default
    \item learning rate decay frequency = default
    \item norm type = instance norm
    \item Adam optimizer momentum parameter = default
    \item pre-processing steps = resize and crop (default) (with loading size and cropping size as default)
\end{itemize}

The results obtained with this configuration can be seen in section \ref{results}.

\vspace{-3mm}
\subsection{StarGAN}
\label{subsection:Implementation:StarGAN}
\subsubsection{Model}
The StarGAN \cite{StarGAN} model has the advantage to train a single generator and discriminator to learn mappings across multiple domains. In our case, it means that the StarGAN architectures could have the capability to learn the translation between demoisaiced raw images to retouched images for each artist but also the translation between artists and from retouched to demoisaiced raw images.

\vspace{-3mm}
\subsubsection{Our usage}
Since we tested multiple GANs and prioritized CycleGAN, the tuning the hyperparameters of StarGAN has not been extensive, mainly because of the time it takes to train a GAN and the available resources to do so. We started with the default configuration proposed in the training section for custom dataset in the readme of official StarGAN GitHub repository: \url{https://github.com/yunjey/StarGAN} and further tried to change the loss and the number of iterations to obtain satisfying results. The script has not been changed; we used the original code found on the GitHub repository.

\vspace{-3mm}
\subsubsection{Training}
We have been able to run 4 training during the available time for this project. The results are described in section \ref{subsection:Results:StarGAN}. Each training has been performed with the dataset resized to 256 pixels on the longest side (see section \ref{subsection:preprocessing}). The first, second and third training have been performed with 27,000 out of the 30,000 images. The first 500 images for each expert and the first 500 images for the raw domain have been kept for testing (10\%). The fourth training has been performed with a reduced dataset of 3,000 images, only the first 500 images of each domain. The training was limited to 2,700 images, the first 50 images for each expert and the first 50 images for the raw domain have been kept for testing (10\%).

The first training has been run with the default initial learning rates (i.e. \verb|g_lr| = \verb|d_lr| = 0.0001), the default number of iterations (i.e. 200,000) and the default batch size (i.e. 16).

The second training has been run with much higher initial learning rates (0.001), the default 200,000 iterations and the default batch size of 16.

The third training has been run with a bigger initial learning rate than the default but smaller than the second training (0.0003), the default 200,000 iterations and the default batch size of 16.

The fourth training has been run with the default learning rates (i.e. \verb|g_lr| = \verb|d_lr| = 0.0001), 1,000,000 iterations and a batch size of 8 to reduce the training time.

\subsection{Pix2Pix}

\label{implementation:pix2pix}
\subsubsection{Model}
Pix2Pix comes from \cite{pix2pix}. The main difference with CycleGAN is that Pix2Pix works only on a paired and aligned dataset.

Pix2Pix uses a conditional generative adversarial network (cGAN), explained in section \ref{section:litterature}, to learn a mapping from an input image to an output image. The structure of the generator is called an ``encoder-decoder". To improve results, the authors of \cite{fivek} modified the generator by using a ``U-Net".

The final objective of this particular model is
\begin{equation}
    G^* = \arg\min_G\max_D\mathcal{L}_{cGAN}(G,D)+\lambda\mathcal{L}_{L1}(G)
\end{equation}
where $L1$ norm was used as it encourages less blurring.

We invite the reader to look at \cite{fivek} for more specific details.

\subsubsection{Our usage}
After some attempts with the original code, we found the results good enough to run with the architecture created by the authors. Our generator is based on a ResNet of 9 blocks, a more recent architecture than the U-Net, and the discriminator is a 70x70 PatchGAN, a kind of convolutional network where each patch is processed identically and independently.
We tried some tweaking of the parameters:
\begin{itemize}
    \itemsep-5pt
    \item the number of epochs with the initial learning rate
    \item the number of epochs to linearly decay learning rate to zero
    \item the pre-processing
\end{itemize}
Where, as in \ref{subsection:CycleGAN}, the learning rate stays the same for a certain number of epochs before decaying for another number of epochs. More of these parameters are explained in the section about the CycleGAN, as the code for both has the same base.

The default values for both types of epochs was 100. While changing the parameters, at first, we noticed that augmenting the number of epochs with normal learning rate (\textit{lr}) and epochs with decaying \textit{lr}, results were getting better. However, later on, when epochs were approaching 500 epochs for normal \textit{lr} and decaying \textit{lr}, the model seemed to overfit as pattern of noises appeared.

Finally, we concluded that the best hyperparameters for the Pix2Pix model are 200 epochs with normal \textit{lr} and 300 epochs with decaying \textit{lr}. Precise results and comparison are shown in the \textit{Results} section below.

\vspace{-3mm}
\section{Results}
\vspace{-3mm}
\label{results}
A specific artist was chosen to compare our results from multiple models and was chosen arbitrarily. Bychkovsky et al. \cite{fivek} had chosen the artist C based on a survey they made. While on our side we did not have the time to do this, we chose to do all our computations mainly on artist D.

\vspace{-3mm}
\subsection{Fr\'echet Inception Distance}
\vspace{-3mm}
The Fr\'echet Inception distance \cite{FID}, also known as FID, is a metric that allows to quantify and score the results of a GAN. It is a way to measure the performance of GANs since the loss function doesn't hold the same signification as in the other Neural Networks. It was formulated two years ago, because Inception Score \cite{Inception_score} had flaws in some dataset configurations. It uses a layer of the Inception neural network that recognizes 2048 features in an image. Passing an image through this layer creates a d-vector of size 2048 with values that represent of much a given feature is present in the image. We then compare the average of all the images in the true image domain B, and the average of all generated image in domain B. These two vectors of the same size represent the two different sets of images, one true and one fake. We then use the formula 
\begin{equation*}
d^2 = ||mu_1 - mu_2||^2 + Tr(C_1 + C_2 - 2 \sqrt{C_1*C_2})
\end{equation*}
which calculates a distance between the two vectors, specifically the Fr\'echet distance also known as Wasserstein-2 distance. 

We use this metric in addition to our visual appreciation to rank performance of our models. 

\vspace{-3mm}
\subsection{Baseline methods}
\vspace{-3mm}
The baseline methods from \ref{subsection:baseline-methods} were used to compute all 5000 raw images from the dataset with artist D as ground truth. The ideal GPR had an $\alpha=e^{-20}$ for all artists. 

The computation of all images took about two hours, a reasonable time compared to all other models described in this paper. 

Displayed below in figures \ref{fig:test_rgb_D} and \ref{fig:gt_rgb_D} 
are artist D 
approximations from our baseline methods model compared to the ground truth. As we can see, colors are not modified but the luminosity of each picture, shadows and bright spots are a lot closer to the artist images.

\begin{figure}[h]
\begin{minipage}{.5\textwidth}
  \centering
  \captionsetup{width=.8\linewidth}
  \includegraphics[width=.7\linewidth]{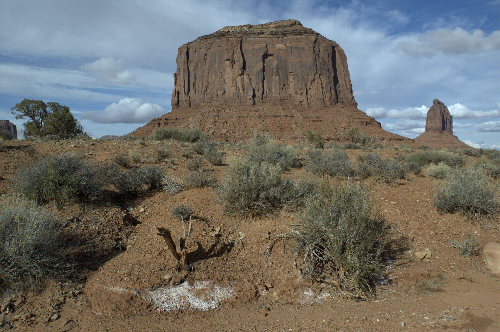}
  \captionof{figure}{Baseline methods model output for artist D.}
  \label{fig:test_rgb_D}
\end{minipage}
\begin{minipage}{.5\textwidth}
  \centering
  \captionsetup{width=.8\linewidth}
  \includegraphics[width=.7\linewidth]{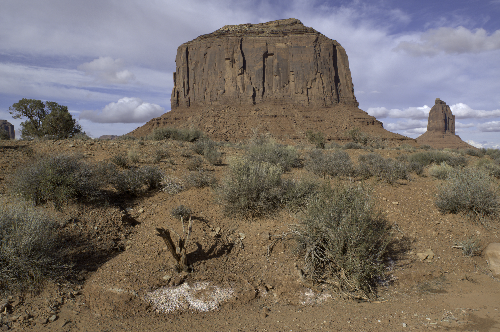}
  \captionof{figure}{Ground truth image for artist D.}
  \label{fig:gt_rgb_D}
\end{minipage}
\vfill
\end{figure}

FID results for artist D was 4.96. This result was surprisingly low. We concluded that the FID score, while giving a quantitative result, did not take much into account look and visual appealing, as they are still differences between the result and ground truth. Also, the output of the baseline models was of a much higher resolution compared to the GANs, which needed very small images to perform efficiently, and the transformation only consisted of a mapping of values.

\vspace{-3mm}
\subsection{CycleGAN}
\vspace{-3mm}
\label{subsection:Results:CycleGAN}

\begin{figure}[h]
\begin{minipage}{.32\textwidth}
  \centering
  \captionsetup{width=.8\linewidth}
  \includegraphics[width=.8\linewidth]{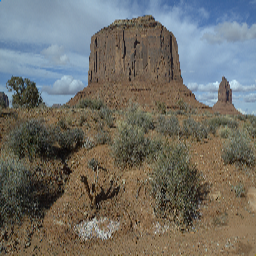}
  \captionof{figure}{Real A}
  \label{fig:adobe_A}
\end{minipage}
\begin{minipage}{.32\textwidth}
  \centering
  \captionsetup{width=.8\linewidth}
  \includegraphics[width=.8\linewidth]{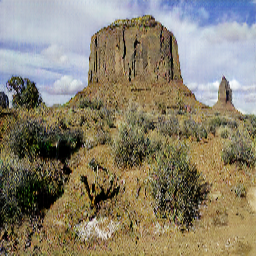}
  \captionof{figure}{Fake B}
  \label{fig:adobe_fB}
\end{minipage}
\begin{minipage}{.32\textwidth}
  \centering
  \captionsetup{width=.8\linewidth}
  \includegraphics[width=.8\linewidth]{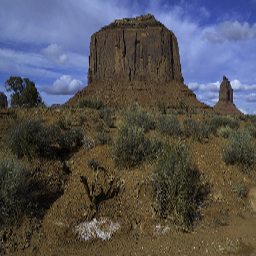}
  \captionof{figure}{Real B}
  \label{fig:adobe_B}
\end{minipage}
\end{figure}

The architecture described in \ref{subsection:CycleGAN} has been run numerous times. As mentioned, we first tried the default parameters with an image size of $256 \times 256$ pixels. For the followings results, we always present the pre-processed RAW image, the generated image in the destination image space, and the ground-truth corresponding. Figures \ref{fig:adobe_A}, \ref{fig:adobe_fB}, \ref{fig:adobe_B} are results with the defaults parameters.

\begin{figure}[!h]
\begin{minipage}{.32\textwidth}
  \centering
  \captionsetup{width=.8\linewidth}
  \includegraphics[width=.8\linewidth]{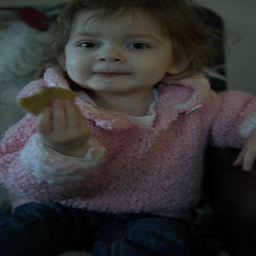}
  \captionof{figure}{Real A}
  \label{fig:vanilla_A2}
\end{minipage}
\begin{minipage}{.32\textwidth}
  \centering
  \captionsetup{width=.8\linewidth}
  \includegraphics[width=.8\linewidth]{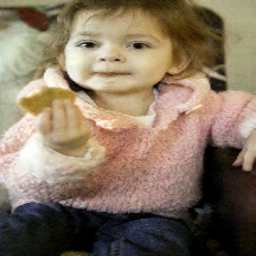}
  \captionof{figure}{Fake B}
  \label{fig:vanilla_fB2}
\end{minipage}
\begin{minipage}{.32\textwidth}
  \centering
  \captionsetup{width=.8\linewidth}
  \includegraphics[width=.8\linewidth]{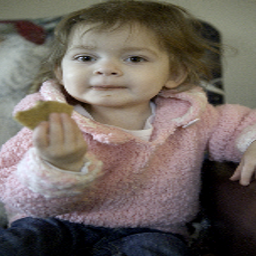}
  \captionof{figure}{Real B}
  \label{fig:vanilla_B2}
\end{minipage}
\end{figure}

We eventually started playing around with hyper-parameters, with some bad results. They are illustrated in section \ref{appendix}.
And finally the configuration that yields the better results, discussed in \ref{subsection:CycleGAN} and illustrated in figures \ref{fig:vanilla_A1}, \ref{fig:vanilla_fB1}, \ref{fig:vanilla_B1}, \ref{fig:vanilla_A2}, \ref{fig:vanilla_fB2} and \ref{fig:vanilla_B2}: 

\begin{figure}[!h]
\begin{minipage}{.32\textwidth}
  \centering
  \captionsetup{width=.8\linewidth}
  \includegraphics[width=.8\linewidth]{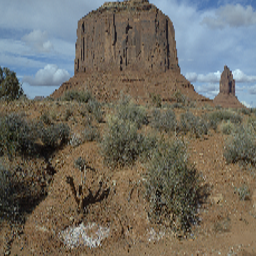}
  \captionof{figure}{Real A}
  \label{fig:vanilla_A1}
\end{minipage}
\begin{minipage}{.32\textwidth}
  \centering
  \captionsetup{width=.8\linewidth}
  \includegraphics[width=.8\linewidth]{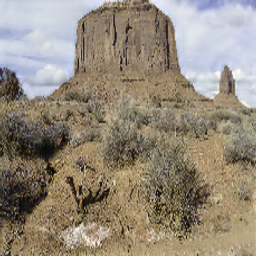}
  \captionof{figure}{Fake B}
  \label{fig:vanilla_fB1}
\end{minipage}
\begin{minipage}{.32\textwidth}
  \centering
  \captionsetup{width=.8\linewidth}
  \includegraphics[width=.8\linewidth]{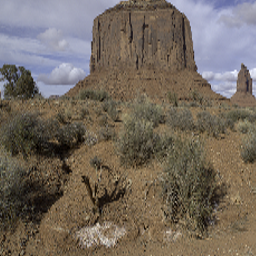}
  \captionof{figure}{Real B}
  \label{fig:vanilla_B1}
\end{minipage}
\end{figure}

As one can see, there are visual differences, but our perception of ``best performing" hyper-parameters is helped by the FID score of the different runs.

\begin{table}[!h]
    \centering
    \begin{tabular}{c|c}
        Model hyperparameters~ & ~FID score \\
        \hline\hline
        random gaussian noise & 483.58\\
        default parameters & 88.23\\
        bad results configuration & 257.54\\
        optimal configuration described in \ref{subsection:CycleGAN} & \textbf{36.82}\\
        \hline
    \end{tabular}
    \vspace{3pt}
    \caption{}
    \label{tab:cyclegan_fid}
\vspace{-10mm}
\end{table}

The FID scores presented in Table \ref{tab:cyclegan_fid} correlate well with the visual impression of the images. The lower the score, the better the image. The distance between random gaussian noise and the ground-truth is also included, for comparison purposes. The iteration with the lower FID score currently represents our best result, both with the metric and visually. We will simply conclude that for our problem, the cross-entropy loss proposed in \cite{original_GAN} seems to work best when coupled with the other default values. That could however change with another dataset. 

\vspace{-3mm}
\subsection{StarGAN}
\vspace{-3mm}
\label{subsection:Results:StarGAN}

The StarGAN architecture has been trained four times. All specifications are provided in section \ref{subsection:Implementation:StarGAN}. In this section we describe and interpret the results of each training and resulting models.

The results are presented in figures with the following format: the first column is the input image. All the presented results where produced with a retouched image of a single expert as input. The second column appeared to be the attempt of the architecture to retrieve the raw image from the input. The third, fourth, fifth and sixth columns are the attempt of the architecture to translate from the input to one of the other artists domain.

\begin{figure}[h]
  \centering
  \captionsetup{width=.8\linewidth}
  \includegraphics[width=.7\linewidth]{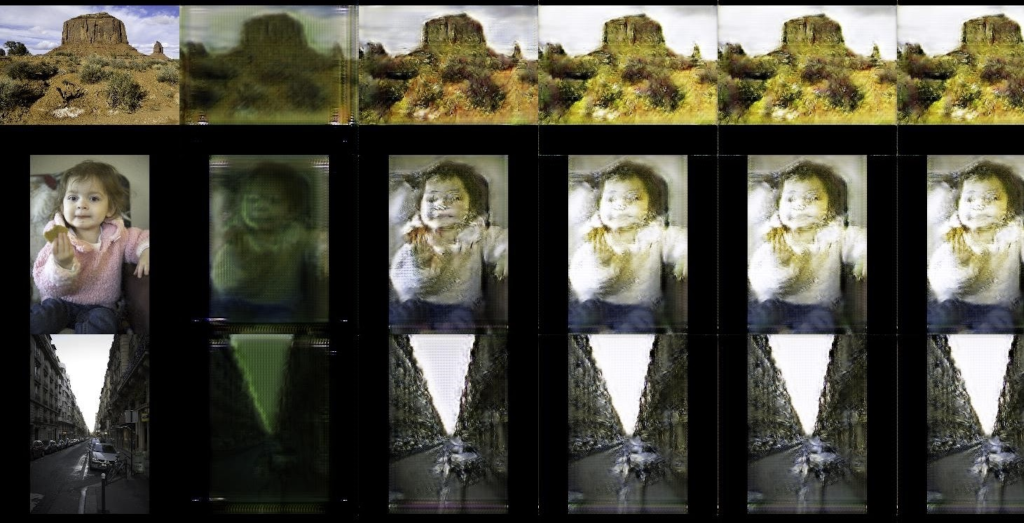}
  \captionof{figure}{Result of the first model obtained with StarGAN.}
  \label{fig:StarGAN_first_run}
\end{figure}

The resulting model of the first training resulted in images as in Figure \ref{fig:StarGAN_first_run}. The result is visually not appealing at all. The model produces a disturbing kind of a blur, distortion in part of the image, such as the face of the child and adds a squared pattern to the images. On the other side, it seems that the architecture managed to learn how to grasp partially the changes in contrast and luminance. The author of StarGAN \cite{StarGAN} designed and tuned this network to work on faces only, which is a much simpler task than natural scenes in general.

\begin{figure}[h]
\begin{minipage}{.5\textwidth}
  \centering
  \captionsetup{width=.8\linewidth}
  \includegraphics[width=.7\linewidth]{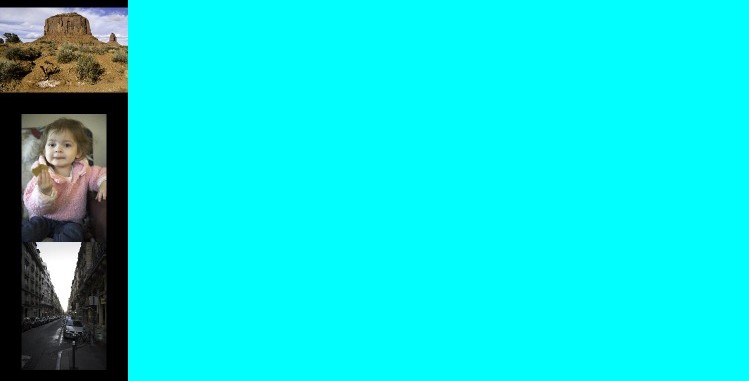}
  \captionof{figure}{Result of the second model obtained with StarGAN.}
  \label{fig:StarGAN_second_run}
\end{minipage}
\begin{minipage}{.5\textwidth}
  \centering
  \captionsetup{width=.8\linewidth}
  \includegraphics[width=.7\linewidth]{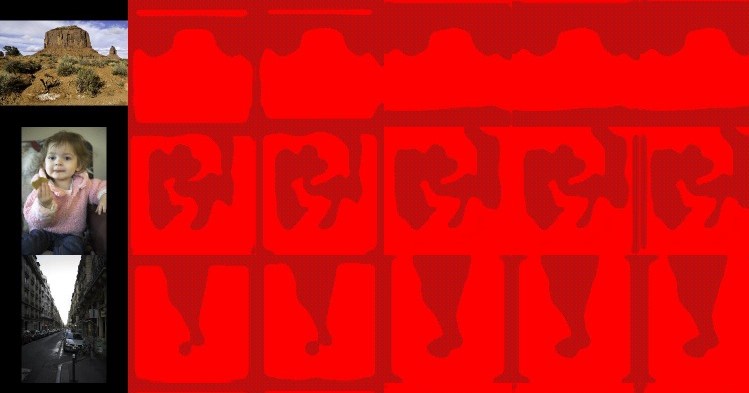}
  \captionof{figure}{Result of the third model obtained with StarGAN.}
  \label{fig:StarGAN_third_run}
\end{minipage}
\end{figure}

Following the previous conclusion we tried a second training with a much higher learning rate (10 times bigger than the default) which was way too big as attested by Figure \ref{fig:StarGAN_second_run}. The learning in this architecture highly diverged and the results were not usable at all. A similar issue arised with the third training result with a learning rate three times the default. An extract of the results is depicted in Figure \ref{fig:StarGAN_third_run}.

\begin{figure}[!h]
  \centering
  \captionsetup{width=.8\linewidth}
  \includegraphics[width=.7\linewidth]{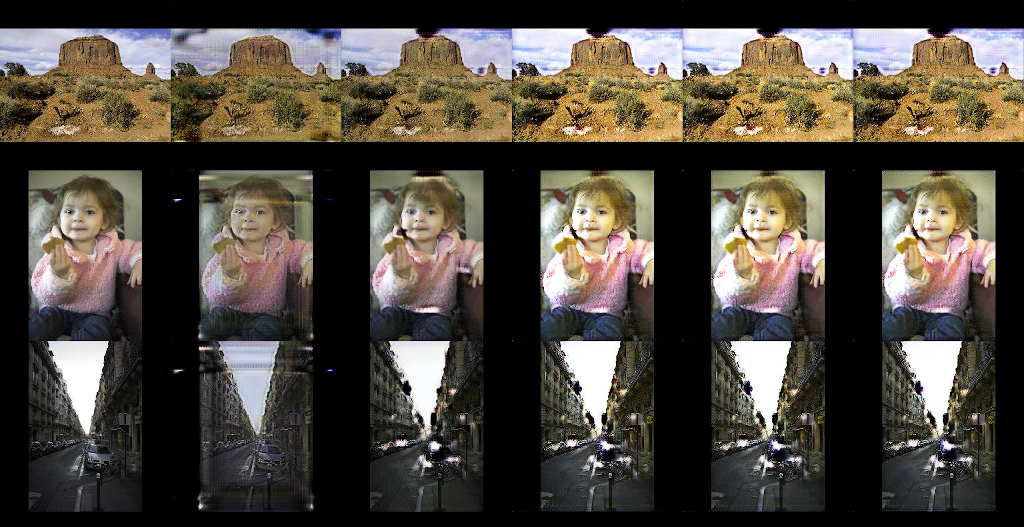}
  \captionof{figure}{Result of the fourth model obtained with StarGAN.}
  \label{fig:StarGAN_fourth_run}
\end{figure}

Based on the result of the second and third model we tried a fourth training with the default learning rate but 1,000,000 iterations to give time for the model to converge. The results where visually significantly better as attested by Figure \ref{fig:StarGAN_fourth_run}. The previously mentioned blur and distortions were significantly reduced but the squared pattern remains and neither the style of the artists nor the demosaiced raw format seems to be reproduce faithfully by the model.

Considering the time remaining for the project and the good results we already obtained with CycleGAN (\ref{subsection:Results:CycleGAN}) and Pix2Pix (\ref{subsection:result_pix2pix}), no further training where performed for StarGAN but it does not exclude that the StarGAN architecture could perform well with proper tuning.

\subsubsection{Note:} The visual result was too bad for the FID score to actually matter, hence it was not computed for the results of section \ref{subsection:Results:StarGAN}.

\newpage
\vspace{-3mm}
\subsection{Pix2Pix}
\vspace{-3mm}
\label{subsection:result_pix2pix}

For Pix2Pix models, a split of our dataset was made to have 4000 images to train, 1000 images to test. As explained in the \textit{Implementation} section about Pix2Pix, the images need to be perfectly aligned, but images dimensions between artist's and raw images were slightly off.\\
Thus, we resized the artist images to having a long edge of 256 pixels, and the raw ones to the exact same dimensions of the artist's. This fixed our problem of alignment but might have caused an imperfect registering of the pixels, meaning that a specific pixel on the raw image, might not be the same on the artist's image, even though they have the same dimensions.

\begin{figure}[!h]
\begin{minipage}{.24\textwidth}
  \centering
  \captionsetup{width=.8\linewidth}
  \includegraphics[width=.9\linewidth]{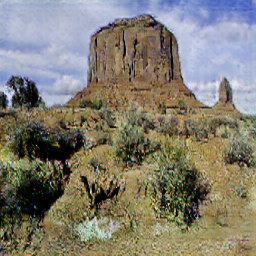}
  \captionof{figure}{Fake C}
  \label{fig:pix2pixtest_rgb_C}
\end{minipage}
\begin{minipage}{.24\textwidth}
  \centering
  \captionsetup{width=.8\linewidth}
  \includegraphics[width=.9\linewidth]{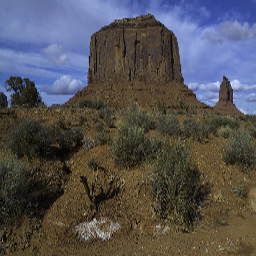}
  \captionof{figure}{Real C}
  \label{fig:pix2pixgt_rgb_C}
\end{minipage}
\begin{minipage}{.24\textwidth}
  \centering
  \captionsetup{width=.8\linewidth}
  \includegraphics[width=.9\linewidth]{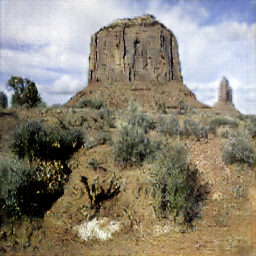}
  \captionof{figure}{Fake D}
  \label{fig:pix2pixtest_rgb_D}
\end{minipage}
\begin{minipage}{.24\textwidth}
  \centering
  \captionsetup{width=.8\linewidth}
  \includegraphics[width=.9\linewidth]{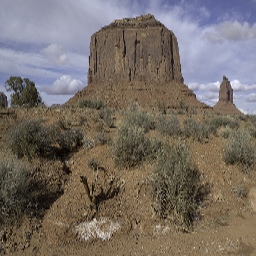}
  \captionof{figure}{Real D}
  \label{fig:pix2pixgt_rgb_D}
\end{minipage}
\end{figure}

While the training of the Pix2Pix models was not too resource-hungry, we were able to make multiple runs, thus giving plenty of evidence.
As we can see in Table \ref{tab:pix2pix_fid} and accordingly in Figure \ref{fig:pix2pixtest_rgb_C} to \ref{fig:pix2pixgt_rgb_D}, the best model was \textit{pix2pixC}.

\textit{pix2pixA} was the first trained model with default parameters. (see \textit{Appendix} \ref{appendix} for images result)\\
\textit{pix2pixB} was the second model trained with number of epochs with inital \textit{lr} equal 500 and number of epochs of decaying \textit{lr} also equal to 500. The resulting images seem blurrier and noisier, we thought of overfitting. (also see \ref{appendix})\\
\textit{pix2pixC} and \textit{pix2pixD} were the models trained with optimal hyperparameters for the number of epochs, 200 normal \textit{lr} and 300 decaying \textit{lr}.

\begin{table}[]
    \centering
    \begin{tabular}{c|c}
        Model name~ & ~FID score \\
        \hline\hline
        pix2pixA & 74.45\\
        pix2pixB & 80.05\\
        pix2pixC & \textbf{63.38}\\
        pix2pixD & 66.05\\
        \hline
    \end{tabular}
    \vspace{3pt}
    \caption{}
    \label{tab:pix2pix_fid}
\vspace{-10mm}
\end{table}

\vspace{-3mm}
\section{Conclusion}
\vspace{-3mm}
 As we've seen, the original idea from the authors of \cite{fivek}, that 90\% of the variance of an image is due to the adjustment of the luminance curve seems about right. The lowest FID was found with this method. Now we need to remember that for GANs, performance is hard to assess. The two standard measures Inception Score \cite{Inception_score} and the Fr\'echet Inception Distance \cite{DBLP:journals/corr/HeuselRUNKH17} give us explicit results, but assessment is often a subjective “beauty contest”.

This is where CycleGAN starts to shine. When compared to the other models, it gives the results that are most pleasing to the eye, with a reasonable FID score. It works well on general images, and not only on specific visage or landscape images.\\
Pix2Pix, based on the same code base, could do as well or better, but has the constraint of having paired and aligned images to work with. This can be hard to bypass for a lambda user that simply wants to apply retouching.\\
StarGAN looked promising but seems to need a lot of resources to have a potential interesting result. It might also be too much for someone to have multiples images to learn transformation to multiple domains.

In the future, it would be interesting to further explore StarGAN and its ability to translate between either a single pair of domains, or multiple ones, and to have a look at new GAN architecture that will eventually come out.

\newpage
\section{Appendix}
\vspace{-3mm}
\label{appendix}

\subsection{Additional results for CycleGAN}
\vspace{-3mm}
For the sake of completion, we provide here examples of the recursive images generated by CycleGAN (the actual cycle part). These correspond to RAW images but generating them is something pretty useless in the scope of our project, hence them not being included in \ref{results}. 

\begin{figure}[!h]
\begin{minipage}{.32\textwidth}
  \centering
  \captionsetup{width=.8\linewidth}
  \includegraphics[width=.8\linewidth]{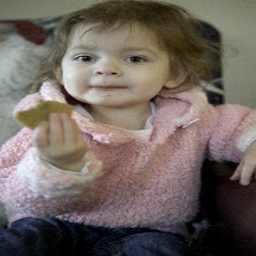}
  \captionof{figure}{Real B}
\end{minipage}
\begin{minipage}{.32\textwidth}
  \centering
  \captionsetup{width=.8\linewidth}
  \includegraphics[width=.8\linewidth]{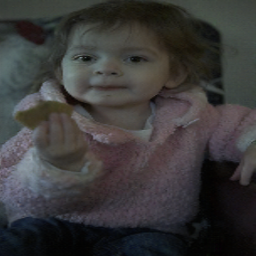}
  \captionof{figure}{Fake A}
\end{minipage}
\begin{minipage}{.32\textwidth}
  \centering
  \captionsetup{width=.8\linewidth}
  \includegraphics[width=.8\linewidth]{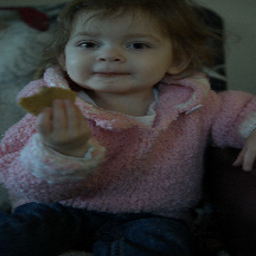}
  \captionof{figure}{Real A}
\end{minipage}
\vspace{-15pt}
\end{figure}

\begin{figure}[h]
\begin{minipage}{.32\textwidth}
  \centering
  \captionsetup{width=.8\linewidth}
  \includegraphics[width=.8\linewidth]{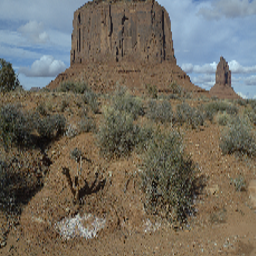}
  \captionof{figure}{Real A}
  \label{fig:artist_A}
\end{minipage}
\begin{minipage}{.32\textwidth}
  \centering
  \captionsetup{width=.8\linewidth}
  \includegraphics[width=.8\linewidth]{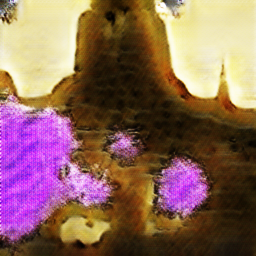}
  \captionof{figure}{Fake B}
  \label{fig:artist_fB}
\end{minipage}
\begin{minipage}{.32\textwidth}
  \centering
  \captionsetup{width=.8\linewidth}
  \includegraphics[width=.8\linewidth]{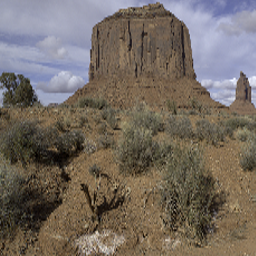}
  \captionof{figure}{Real B}
  \label{fig:artist_B}
\end{minipage}
\vspace{-20pt}
\end{figure}

CycleGAN also had some bad runs, as illustrated in figures \ref{fig:artist_A}, \ref{fig:artist_fB} and \ref{fig:artist_B}. 

\vspace{-3mm}
\subsection{Additional results for Pix2Pix}
\vspace{-3mm}

Here are two more images examples for the \textit{pix2pixA} and \textit{pix2pixB} models.

\begin{figure}[!h]
\begin{minipage}{.24\textwidth}
  \centering
  \captionsetup{width=.8\linewidth}
  \includegraphics[width=.9\linewidth]{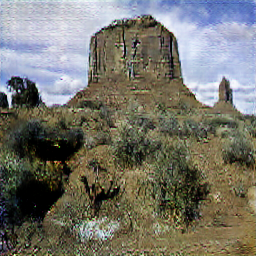}
  \captionof{figure}{Fake A}
  \label{fig:test_rgb_A}
\end{minipage}
\begin{minipage}{.24\textwidth}
  \centering
  \captionsetup{width=.8\linewidth}
  \includegraphics[width=.9\linewidth]{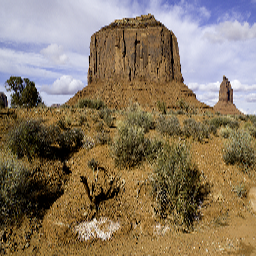}
  \captionof{figure}{Real A}
  \label{fig:gt_rgb_A}
\end{minipage}
\begin{minipage}{.24\textwidth}
  \centering
  \captionsetup{width=.8\linewidth}
  \includegraphics[width=.9\linewidth]{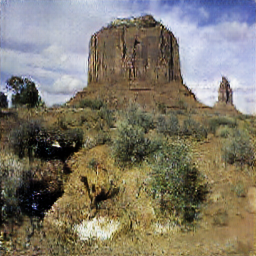}
  \captionof{figure}{Fake B}
  \label{fig:test_rgb_B}
\end{minipage}
\begin{minipage}{.24\textwidth}
  \centering
  \captionsetup{width=.8\linewidth}
  \includegraphics[width=.9\linewidth]{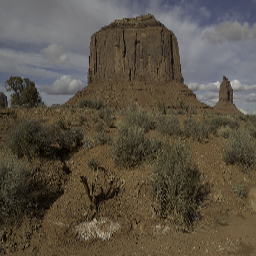}
  \captionof{figure}{Real B}
  \label{fig:gt_rgb_B}
\end{minipage}
\end{figure}

\vspace{-3mm}
\subsection{Results reproduction}
\vspace{-3mm}

All code and results can be reproduced by going to our GitHub repository and following at : \url{https://github.com/MarcBickel/CS-413}

\bibliographystyle{splncs04}
\bibliography{references.bib}

\end{document}